\documentclass[twoside,11pt]{article}

\usepackage{jmlr2e}
\usepackage[utf8]{inputenc}
\usepackage{hyperref}
\usepackage{listings}
\usepackage{booktabs}
\usepackage{amssymb}
\usepackage{multirow}
\usepackage{dblfloatfix}    
\usepackage{csvsimple}
\usepackage{adjustbox}

\def\rot{\rotatebox}

\jmlrheading{1}{2016}{1-15}{8/93;}{9/93}{14-115}{Piotr Szymański and Tomasz Kajdanowicz}

\ShortHeadings{A Python library for Multi-Label Classification}{Piotr Szymański and Tomasz Kajdanowicz}
\firstpageno{1}

\begin{document}

\title{\href{http://scikit-multilearn.github.io}{scikit-multilearn}: A scikit-based Python environment for performing multi-label classification}

\author{\name Piotr Szymański \email piotr.szymanski@\{\href{mailto:piotr.szymanski@pwr.edu.pl}{pwr.edu.pl},\href{mailto:piotr.szymanski@illimites.edu.pl}{illimites.edu.pl}\} \\
       \addr Department of Computational Intelligence \\
       Wrocław University of Science and Technology \\
       Wrocław, Poland\\ \\
       \addr  illimites foundation \\
       Wrocław, Poland 
       \AND
       \name Tomasz Kajdanowicz \email tomasz.kajdanowicz@pwr.edu.pl \\
       \addr Department of Computational Intelligence \\
       Wrocław University of Science and Technology \\
       Wrocław, Poland
}

\editor{Leslie Pack Kaelbling}

\maketitle

\begin{abstract}
    scikit-multilearn is a Python library for performing multi-label classification. The library is compatible with the scikit/scipy ecosystem and uses sparse matrices for all internal operations. It provides native Python implementations of popular multi-label classification methods alongside a novel framework for label space partitioning and division. It includes modern algorithm adaptation methods, network-based label space division approaches, which extracts label dependency information and multi-label embedding classifiers. It provides python wrapped access to the extensive multi-label method stack from Java libraries and makes it possible to extend deep learning single-label methods for multi-label tasks. The library allows multi-label stratification and data set management. The implementation is more efficient in problem transformation than other established libraries, has good test coverage and follows PEP8. Source code and documentation can be downloaded from \url{http://scikit.ml} and also via \texttt{pip}. The library follows BSD licensing scheme. 
\end{abstract}


\section{Introduction}

The Python language with its machine learning library stack has grown to become one of the leading technologies of building models for the industry and developing new methods for the researchers. While the python community boasts with the excellent culture of development, well-defined API traditions and well-performing implementations of methods from most machine learning areas it did not have a high-quality solution for multi-label classification.

In this paper we introduce scikit-multilearn, a well-tested, multi-platform, Python 3 compatible, BSD-licenced library with bleeding edge approaches for multi-label problems. Scikit-multilearn supports state-of-the-art approaches, is actively developed and has a growing user base. We also show that it is faster than its competition in other languages. Scikit-multilearn is also entirely compatible with the scientific/machine learning ecosystem of well-known python libraries which allows it to fill the niche of the missing multi-label classification solution, while also benefits from complementary solutions present in the scikit-learn community, while providing it with efficient implementations of missing classification methods, data stratification, dataset access and manipulation.

We start with a summary of what multi-label classification is in Section \ref{sec:mlc}. Followed by Section \ref{sec:rw} where we present related work and a description of the machine learning ecosystem that scikit-multilearn fits in. In Section \ref{sec:meth} we describe what scikit-multilearn brings to the community. We evaluate how scikit-multilearn performs in comparison to libraries implemented in other languages in Section \ref{sec:bench} and present conclusions and future ideas in Section \ref{sec:conc}.

\section{Multi-label classification\label{sec:mlc}}

Multi-label classification deals with the problem of assigning a subset of available labels to a given observation. Such problems appear in multiple domains - article and website classification, multimedia annotation, music categorization, the discovery of genomics functionalities and are performed either by transforming a problem into a single/multi-class classification problem or by adapting a single/multi-class method to take multi-label information into account. An excellent introduction to the field has been provided by \citet{tsoumakas2009mining}.

\citet{madjarov_extensive_2012} divide approaches multi-label classification into three groups: method adaptation, problem transformation and ensembles thereof. The first idea is to adapt single-label methods to multi-label classification by modifying label assigning fragments in a single-label method. An example of this is introducing a multi-label version of a decision function in Decision Trees \citep[ex.][]{kocev2013tree}.

Problem transformation concentrates on converting the multi-label problem to one or more single-label problems. The most prominent examples include classifying each label separately such as Binary Relevance or Classifier Chains \citep{read2009classifier} and treating each label combination as a separate class in one multi-class problem as in the Label Powerset case. 

Both of these approaches suffer not only from standard problems of machine learning such as over- and under-fitting but also from the issue of label imbalance \citep{sun2009classification} or numerical anomalies related to label ordering when Bayesian approaches are used for taking correlations into account. Ensemble methods aim to correct this by learning multiple classifiers trained on label subspaces \citep[ex. RAkEL by][]{tsoumakas2011random}, observation subsets with pruning and replacement, or analyzing various label orderings in chains \citep[ex. Probabilistic Classifier Chains by][]{dembczynski2010bayes}. 

Multi-label embedding techniques emerged as a response the need to cope with a large label space; these include label space dimensionality reduction techniques that turned Most multi-label embedding methods turn multi-label classification into multivariate regression problem followed by a rule-based or classifier-based correction step. Embedding methods also vary by the principle of how the embedding is performed, these include: Principle Label Space Transformation \citep{tai2012multilabel} based on Principal Component Analysis; Conditional Principal Label Space Transformation \citep{chen2012feature} with Canonical Component Analysis; Feature-aware Implicit Label Space Encoding  \citep{lin2014multi} based on matrix decompositions; SLEEC  \citep{bhatia2015sparse} which uses k-means clustering and per cluster embedding; CLEMS \citep{huang2017cost} which performs Multi-Dimensional Scaling \citep[see][]{kruskal1964multidimensional} with extra multi-label quality measure optimization; and LNEMLC \citep{1812.02956} which embeds label networks using network embeddings such as LINE or node2vec. Most of these use linear, ridge or random forest regressors to predict embeddings for unseen samples and a variant of kNN to classify them with labels. CLEMS and LNEMLC are among the best performing multi-label embeddings at this time.

Multi-label advances in recent years also include developments in a sub-area called extreme multi-label where new methods emerged such as deep-learning based domain-specific approaches for: image classification frameworks \citep{zhao2015deep,wang2016cnn,wei2016hcp} or text \citep{liu2017deep, yen2017ppdsparse}. The field also includes tree-based \citep{prabhu2014fastxml} or embedding-based \citep{bhatia2015sparse} approaches. While extreme multi-label is an interesting task it differs strongly from classical multi-label classification in performance expectations, benchmark datasets and quality measures and falls beyond the main focus of scikit-multilearn but its methods can be used to classify these scale of problems.

\section{Related work\label{sec:rw}}
As with every applied science, research requires environments for performing experiments. The most prominent multi-label classification stack to date (regarding method count and popularity) is implemented in Java: MULAN \citep{mulan} and MEKA: \citep{MEKA}. Both depend heavily on the famous WEKA library \citep{hall_weka_2009} which implements a plethora of classification and regression methods for single-label classification. MULAN and MEKA are large multi-purpose libraries which support not only multi-label classification tasks but also regression and multi-instance/multi-output tasks.

Python's scientific ecosystem's philosophy \citep{oliphant2007python} is a different one, instead of having few large multi-purpose libraries, it provides a foundation of libraries that deliver most important approaches and a network of smaller and more specialized libraries that interact thanks to a well-defined API, coding and documentation standards set out for common machine learning tasks such as prediction or data transformation. With numpy \citep{numpy}, scipy \citep{scipy} as numerical operation and data structure foundations. scikit-learn \citep{scikit}  provides a machine-learning foundation with a different approach than WEKA as it does not aim at being a large mono-library where all functionality is gathered. Instead, it concentrates on implementing most cited well-established methods alongside a stable API design. 

Scikit-learn provides rudimentary multi-label methods however without support for sparse representations of label matrices apart from the multi output classifier: algorithm adaptation approaches such as k nearest neighbors, CART trees and random forests and perceptrons; problem transformation methods Binary Relevance, Classifier Chains and one-vs-rest or one-vs-all usage of multi-class classifiers such as Support Vector Machines \citep{hearst1998support}. However all of these are the most rudimentary variants of listed methods and have been superseded by more modern approaches. Scikit-learn's kNN approach has been improved by MLkNN \citep{mlknn}, deep learning is used much more often than perceptrons, and new tree-based classifiers like fastXML \citep{prabhu2014fastxml} outperform traditional random forests of multi-label trees. One-vs-all/rest multi-class classifiers can now be replaced with inherently multi-label variants of the methods, ex. SVM classification scheme has been enhanced by multi-label SVM approaches such as MLTSVM \citep{chen2016mltsvm}. Binary Relevance provides worse performance than data-driven label space division \citep{szymanski2016data} or stacking \citep{montanes2014dependent}. Classifier chains have been extended in multiple ways to overcome artifacts related to sample distribution changes between label orderings either by learning shorter chains via data-driven label space divisions or chain sampling methods \citep[PCC][]{dembczynski2010bayes}, \citep[MCC][]{read2013efficient}. However scikit-learn's classifiers are extremely useful when used as a base for more comprehensive methods in scikit-mutlilearn such as label space partitioning ensembles or multi-label embedding.

Scikit-learn is also the hub of a large network of complementary libraries for more specific tasks, new techniques or emergent sub-fields. Such libraries include among many others scikit-multiflow and imbalanced-learn.

scikit-multiflow by \citet{skmultiflow} is a multi-label stream library, which differs from the traditional multi-label setting where a method is trained once to predict data - in multi-label streams. Then methods are retrained or adopted as new samples arrive. Even though scikit-multiflow provides reimplementations selected methods of scikit-learn, the authors note that scikit-learn's implementations should be applied while using the library. Authors claim that the provided implementations are less optimal (kNN) or masks over the original scikit-learn code for internal purposes. more Modern version of classifier chains (PCC, MCC) and stream classifiers are the core contribution of the library. They are optimized towards partial updating as new evidence arrives. The library is well unit-tested and should be used for multi-label stream tasks which scikit-multilearn does not handle out of the box. However, scikit-multilearn provides more vibrant and various offering of state of the art methods optimized for the classic multi-label problem formulation.  

imbalanced-learn by \citet{JMLR:v18:16-365} is dedicated to overcoming class imbalance problems in single-label tasks by a variety of under/over-sampling strategies. As many problem transformation approaches transform the problem to a multi-class problem imbalanced-learn can be used to improve learning capabilities of scikit-multilearn classifiers by fitting them to resampled data.

\begin{table}
    \centering
\begin{tabular}{lcccccc}
\toprule
                              &   \multicolumn{2}{c}{Java}  & \multicolumn{2}{c}{Python} &        R & Matlab  \\
                              &   \multicolumn{2}{c}{ }  & \multicolumn{2}{c}{ } &         & Octave  \\
\midrule
                                        &  \rot{90}{MULAN} &     \rot{90}{MEKA} &    \rot{90}{scikit-learn} &    \rot{90}{scikit-multilearn} &    \rot{90}{util.ml} &     \rot{90}{MLC\_toolbox} \\
     \textbf{Algorithm-adaptation methods} &        &        &               &                    &          &               \\
                                       Multi-label SVM &       &       &              & \checkmark                   &          &        \checkmark        \\
                                  Trees &      \checkmark &      \checkmark &             \checkmark &                    &          &              \checkmark  \\
                      Nearest Neighbors &      \checkmark &      \checkmark &             &                  \checkmark &           &              \checkmark  \\
                                 Neural &      \checkmark &      \checkmark &             \checkmark &                  \checkmark &          &               \checkmark \\
\midrule
      \textbf{Problem transformation approaches} &        &        &               &                    &          &               \\
                        To single-class &      \checkmark &      \checkmark &             \checkmark &                  \checkmark &        \checkmark &             \checkmark \\
                         To multi-class &      \checkmark &      \checkmark &               &                  \checkmark &          &             \checkmark \\
                         To deep learning models &       &       &               &                  \checkmark &          &              \\
\midrule
                    \textbf{Ensemble approaches} &        &        &               &                    &          &               \\
                               Stacking &      \checkmark &      \checkmark &             \checkmark &                  \checkmark &        \checkmark &             \checkmark \\
               Label space partitioning &      \checkmark &      \checkmark &               &                  \checkmark &          &             \checkmark \\
                     Overlapping models &      \checkmark &      \checkmark &               &                  \checkmark &          &             \checkmark \\
                             Embeddings &        &        &               &                  \checkmark &          &             \checkmark \\
\midrule
                        \textbf{Related tasks}  &        &        &               &                    &          &               \\
             Multi-label stratification &      \checkmark &      \checkmark &               &                  \checkmark &          &               \\
                       Quality measures &      \checkmark &      \checkmark &             \checkmark &                  P &        \checkmark & \checkmark              \\
           Multi-label dataset management &      \checkmark &      \checkmark &             P &                  \checkmark &        \checkmark &               \\
                             Multi-label regression &      \checkmark &      \checkmark &             \checkmark &                    &          &               \\
                           Multi-output learning &      \checkmark &        &               &                    &          &               \\
                           Sparse label space representation &      &        &  P             & \checkmark                    &          &               \\
               Multi-label ARFF support &      \checkmark &      \checkmark &               &                  \checkmark &        \checkmark &               \\
\midrule
                        \textbf{Project quality} &        &        &               &                    &          &               \\
                 End - to end use cases &        &        &             \checkmark &                  \checkmark &          &               \\
                 
              Developer documentation &    \checkmark    &        &             \checkmark &                  \checkmark &         &               \\
                      API documentation &    \checkmark    &   \checkmark     &             \checkmark &                  \checkmark &  \checkmark        &               \\
                             Unit-tests &      \checkmark &      \checkmark &             \checkmark &                  \checkmark &        \checkmark &               \\
                         Multi-platform &      \checkmark &      \checkmark &             \checkmark &                  \checkmark &        \checkmark &             P \\
 Continous integration on all platforms &        &        &             \checkmark &                  \checkmark &          &               \\
\bottomrule
\end{tabular}
    \caption{Comparison of multi-label classification libraries and their features. Letter P denotes partial functionality: scikit-learn can generate multi-label datasets, but does not offer methods to lead them and manipulate them; sparse support is implemented only for Binary Relevance (MultiOutputClassifier). MLC toolbox offers some of its classifiers Windows-only. scikit-multilearn is compatible with scikit-learn's quality measures. \label{tab:mlc.feature.comparison}}
\end{table}

Domain and sub-field related deep neural networks with multi-label classification support are also available such as magpie\footnote{\url{https://github.com/inspirehep/magpie}}, cnn-rnn for images or a variety of different models for biological data\footnote{\url{https://qdata.github.io/deep4biomed-web/}} for text classification.

Multi-label classification problems can also be addressed in general purse deep learning libraries such as Tensorflow \citep{tensorflow2015} or keras \citep{chollet2015keras}. However, deep neural networks often do not demonstrate an appropriate level of robustness and require significant differences in architecture to accommodate a concrete problem. This introduces an additional level of complication while applying deep learning to multi-label problems. While these libraries are instrumental in their areas, scikit-multilearn is dedicated to providing domain-independent solutions as efficient per domain specialization is rarely possible in one unified code base. However, scikit-multilearn does provide support for using keras-compatible models for multi-label classification independent of the backend used by keras (i.e. tensorflow, cntk) or pytorch (via the skorch\footnote{\url{https://github.com/dnouri/skorch}} library).

Scikit-multilearn fits in the Python community similarly as to how MEKA or MULAN fit to the Java machine learning projects network, however, it is focused on multi-label classification problems to provide state of the art approaches and efficient implementation of more advanced methods.

Recently multi-label classification libraries were made available in R  \citep{R} and Matlab/Octave \citep{octave}. \citet{RJ-2018-041} release the utiml R library built on top of mldr dataset management library \citep{charte-charte:2015} and offers a larger variety of available methods than other R libraries and a more welcoming API. \citet{kimura2017mlc} made a mutli-label toolbox available for Matlab/Octave communities. 

Table \ref{tab:mlc.feature.comparison} provides a general point of view on the functionality offered by multi-label classification methods in Java, Python, R and Matlab/Octave. 

\section{The scikit-multilearn library\label{sec:meth}}

Scikit-multilearn follows the idea of building a multi-label classification library on top of an existing classification solution. In the case of Python, the obvious choice for a base library is the scipy stack with scikit-learn. The concept of scikit-learn compatible projects (scikits in short, not to mistake with the old scipy notion of scikits), has been present for several years in many formats, most prominently exemplified in scikit-contrib. We follow the ideas of these communities, the scikit-learn API principles and licensing. 

The primary goal of scikit-multilearn is to provide an efficient Python implementation of as many multi-label classification algorithms as possible both to the open source community and commercial users of the Python data science stack. With such focus in mind, we concentrate on delivering a library dedicated to provide domain-independent solutions for solving multi-label classification problems, allowing other toolkits to excel in their areas such as: scikit-multiflow when it comes to multi-label streams or imbalanced-learn when it comes to handling imbalance multi-class data or many deep learning approaches described in the previous section. 

\paragraph{Extensions of scikit-learn.}
Scikit-multilearn extends the multi-label classification offering of scikit-learn by:
\begin{itemize}
    \item extending the family of classifiers available to the Python community with more advanced algorithm adaptation and problem transformation (label space division) approaches
    \item implementing multi-label embedding-based classification methods
    \item providing additional features too specific for scikit-learn such as the general framework for classification based on label space division or the wrapper for MEKA,
    \item implementing non-classification tools for multi-label classification, including multi-label dataset manipulation from ARFF, loading and saving sparse representations of these datasets and the only python implementations of multi-label data stratification methods (see work by \citep{sechidis2011stratification}, \citep{szymanski2017network})
\end{itemize}

\paragraph{Advanced Multi-label Adapted Algorithms.}
Scikit-multilearn provides implementations of recent multi-label algorithm adaptation methods, but do not meet the selectiveness criteria of scikit-learn due to their novelty. These include: 
\begin{itemize}
\item ML-ARAM: the hierarchical multi-label adaptive resonance associative map \citep{mlaram}
\item BRkNN and MLkNN: Binary Relevance kNN \citep{brknn} and multi-label kNN \citep{mlknn} - their scikit-multilearn implementations are built on top of scikit-learn's NearestNeighbors multiclass classifier,
\item MLTSVM: the multi-label twin support vector machine \citep{chen2016mltsvm} which does not require a training quadratic number of SVM classifiers.
\end{itemize}

\paragraph{Problem Transformation with Label Space Division.}
One of the significant contributions of the library is a framework for performing ensemble classification with different label space division strategies. The library provides ways to divide the label space based on:
\begin{itemize}
    \item clustering of the label matrix representations, using any of the scikit-learn compatible clusterers (i.e. classes that inherit scikit-learn's \texttt{ClusterMixin} base class). The clusterer is executed on a transposed label assignment matrix and detected clustering is used as the label space division for the ensemble method. An example of this approach is the k-means clustering of labels which has been applied by many approaches \citep{tsoumakas2008effective,yu2017classifier}).
    \item communities detected in a network which embeds information about relations between labels. Scikit-multilearn is a general API for defining graph generators based on the label matrix, such graphs can exploit relations based on co-occurrence, correlation or any other pairwise measure defined over labels. The library supports community detection methods based on three popular graph/network packages in Python:  \citep[NetworkX, BSD by][]{hagberg2008exploring}), \citep[igraph, GPL by][]{igraph}, \citep[graphtool, GPL by][]{peixoto_graph}. An example of this approach is the label space division based on label co-occurrence graphs \citep{e18080282}.
    \item random division, present in the random k-label sets (RAkEL) approaches
    \item predefined fixed division obtained from expert knowledge
\end{itemize}

Scikit-multilearn provides a partitioning and a voting majority ensemble approach, where in both a sub-classifier is trained on every label subset from a label space partition. In the partitioning variant, the predicted results are the union of results from each of the classifiers. In the voting ensemble, the label is assigned if the majority of classifiers trained on a label subspace which contain the label have assigned it to a given sample.

\paragraph{Multi-Label Embeddings.} Another significant contribution of the library is a framework for multi-label classification using multi-label embeddings. Scikit-multilearn is the first Python library to provide a general multi-label embedding scheme with an embedding regressor and embedding-based classifier. Currently two state of the art embedding approaches are provided Cost-Sensitive Label Embedding with Multidimensional Scaling (CLEMS)\citep{huang2017cost}, Label Network Embeddings for Multi-Label Classification (LNEMLC)\citep{1812.02956} alongside a general embedder that can use any scikit-learn manifold learning or dimension reduction to embed the output space.

\paragraph{Compatibility.}
scikit-multilearn aims to be compatible with the Python data science stack. It follows the scikit API and requirements specified in check\_estimator code. Multi-label classifiers in scikit-multilearn inherit scikit-learn's \texttt{BaseEstimator} and \texttt{Classifier}\allowbreak\texttt{Mixin} classes. They can be thus easily incorporated into scikit pipelines and cross-validations, evaluation measures, and feature space transformators. It is easy to use scikit-learn's extensive feature-space manipulation methods, single-label classifiers and classification evaluation functions with scikit-multilearn.

\paragraph{BSD licensing.}
As innovations in machine learning happen in both academia and companies, it is important to create a library that allows both communities to grow and profit from common work. Thus, scikit-multilearn is released under the BSD license to permit the for-profit ecosystem to use the library while opening the possibility of sharing development and maintenance tasks. While most of scikit-multilearn's functionality is available via BSD licensed libraries, the package also allows using GPL-licensed libraries to be used for label space division such as igraph or graphtool. Importing from \texttt{skmultilearn.\allowbreak cluster.igraph} or \texttt{skmultilearn.\allowbreak cluster.graphtool} will require your code to submit to GPL licensing requirements.

\paragraph{Access to the Java stack.} 
The library provides scikit-compatible wrapper to reference libraries such as MEKA, MULAN and WEKA through MEKA when a need arises to use methods that have not yet been implemented in Python natively. Using these libraries via scikit-multilearn wrapper does not induce a licensing (GPL-BSD) conflict. Support for MULAN is limited to methods available via MEKA's \texttt{meka.\allowbreak classifiers.\allowbreak multilabel.\allowbreak MULAN} classifier\footnote{\url{http://meka.sourceforge.net/api-1.9/meka/classifiers/multilabel/MULAN.html\#setMethod-java.lang.String-}}.

\paragraph{Using Deep Learning models.} Scikit-multilearn provides a wrapper that allows using any Keras \citep{chollet2015keras} compatible backend such as tensorflow \citep{tensorflow2015} CNTK \citep{Seide:2016:CMO:2939672.2945397} or pytorch \citep{paszke2017automatic}, to provide a single-class or multi-class model that can be used to solve multi-label problems through problem transformation approaches.

\paragraph{Sparsity.}
Multi-label classification problems often do not provide cases for all possible label combinations - in most benchmark sets there are less than 5\% labels per row on average. The output space is thus very sparse. As opposed to MULAN and MEKA, scikit-multilearn operates on sparse matrices internally yielding a large memory boost as displayed in Figure \ref{fig:benchmark}.

\paragraph{Using scikit-multilearn.} The steps for using scikit-multilearn are very simple: loading the data, selecting the method and performing classification. The library supports loading matrices from all well-established formats thanks to scipy and operates on scipy/numpy representations internally, converting to dense matrices or lists of lists only if required by a scikit-base classifier, if one is employed in the problem transformation scenario.

\paragraph{Other output types and subproblems.} Scikit-multilearn does not support multi-label regression or multi-output prediction at the moment. Multi-label regression is considered for future releases, while multi-output approaches will not be supported due to scikit-multilearn's concentration on efficient sparse matrix usage throughout the codebase. Extreme Multi-label Classification is not the main focus of scikit-multilearn but both the label network embedding methods and label space partitioning methods are capable of handling extreme multi-label datasets. Keras-based deep learning models can also be deployed with the label partitioning strategy to solve extreme multi-label problems.

\paragraph{Structure of scikit-multilearn.} Scikit-multilearn provides three subpackages following the established categorization of multi-label methods: method adaptation approach (in \texttt{skmultilearn.adapt}), problem transformation approach (in \texttt{skmultilearn.\allowbreak pt}) and ensembles of the two (in \texttt{skmultilearn.\allowbreak ensemble}). The label space division methods and label relationship graph builders are present in \texttt{skmultilearn.\allowbreak cluster}. To use multi-label-adapted version of a single-label method one only needs to import it and instantiate an object of the class. A problem transformation approach takes a scikit-learn compatible base classifier and optional information whether the base classifier supports sparse input. Ensemble methods require a classifier and relevant method's parameters. The classifier is trained using the fit method, which takes the training input and output matrices, and classification is performed using the predict method on test observations - just as in scikit-learn. The Meka wrapper provides a scikit compatible classifier class in \texttt{skmultilearn.meka}. Dataset download, loading and manipulation tools are available in \texttt{skmultilearn.\allowbreak dataset}. Multi-label data stratification methods and dataset fold division quality measures are available in \texttt{skmultilearn.\allowbreak model\_selection}. Multi-label embedding approaches are located in \texttt{skmultilearn.\allowbreak embeddings}.

\subsection{Documentation and Availability} Scikit-multilearn is hosted on Github and managed via the fork and pull-request paradigm. Both user and developer documentation is available at \url{http://scikit.ml/}. After 4 years of development, in July 2018, the library has reached a stable 0.1.0 release. The newest version - 0.2.0 - was released in December 2018. The code is released under BSD licence using Github and releases are available via PyPi. Implementation follows PEP8 and is maintained with a high level of test coverage (82\% for the 0.2.0 release). The development undergoes continuous integration on Windows, Ubuntu Linux and MacOS X, with Python 2.7 and Python 3.3. At the moment of the 0.2.0 release scikit-multilearn has been starred 242 times and forked 72 times. Development is managed on the Github repository \texttt{scikit-multilearn}\footnote{\url{https://github.com/scikit-multilearn/scikit-multilearn}} and is accompanied with communication and discussion in the slack channel\footnote{\url{https://scikit-ml.slack.com}}. All commits undergo continuous testing on Windows, Ubuntu and MacOSX, both under Python 2.7 and 3.3. The library also has its tag\footnote{\url{https://stackoverflow.com/tags/scikit-multilearn/info}} on StackOverflow.

\section{Benchmark\label{sec:bench}}
Scikit-multilearn is faster than its competition. We have tested MEKA, MULAN and scikit-multilearn on 12 well-cited benchmark multi-label classification datasets using two comparison scenarios: Binary Relevance and Label Powerset. The two selected problem transformation approaches are widely used both in regular classification tasks and as the base for more sophisticated methods. We did not test algorithm adaptation methods as there are no algorithm adaptation methods present in all three libraries. We use these methods to illustrate two aspects of the classification performance of the libraries: the cost of using many classifiers with splitting operations performed on the label space matrix and the cost of using a single classifier which requires to access all label combinations to perform the transformation. To minimize the impact of base classifiers, we have decided to use a fast Random Forest base classifier with 10 trees. As Octave does not provide Matlab's random forest implementation, we used the one provided by the shogun toolbox \citep{soeren_sonnenburg_2017_1067840}. We have checked the classification quality and did not find significant differences between Hamming Loss, Jaccard and Accuracy scores between the outputs. The benchmark is performed with scikit-multilearn 0.1.0, MEKA 1.9.2, MULAN 1.5.0, scikit-learn 0.19.2, Octave 4.2.2, shogun 6.1.3, MLC Toolbox for Matlab/Octave code from Github master commit hash \texttt{e798779}, R 3.4.1, utiml 0.1.2. As utiml does not provide a Label Powerset implementation, we od not evaluate it in this subexperiment.

\begin{figure}[t]
  \includegraphics[width=1\textwidth]{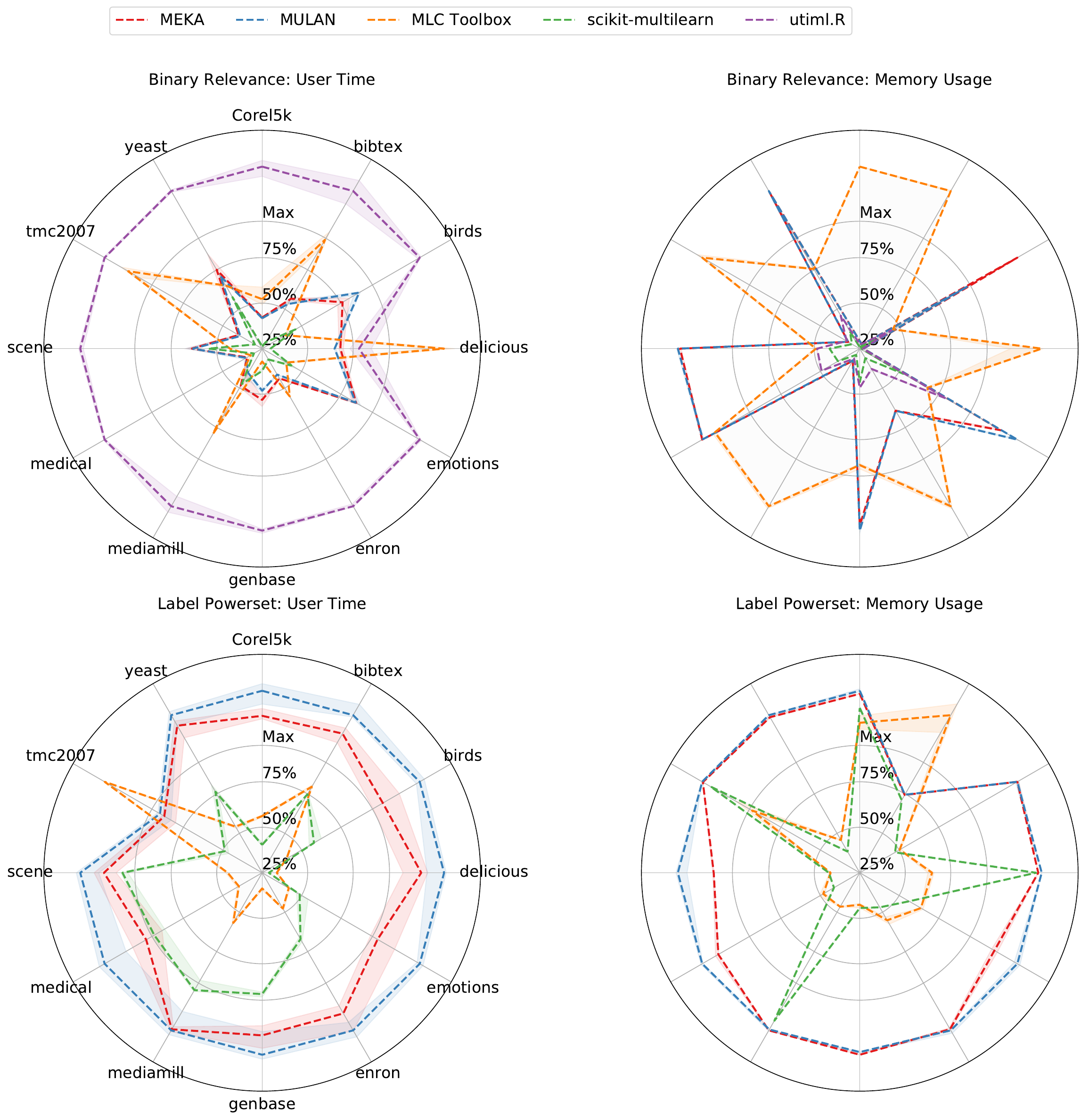}
  \caption{User running time (s) and memory usage ranks of utiml, MLC Toolbox, scikit-multilearn, Meka and Mulan, with RandomForest, and two different multi-label classifiers.\label{fig:benchmark}}
\end{figure}

Figure \ref{fig:benchmark} presents the time and memory required to perform classification, including loading the ARFF dataset (or matlab export in case of the toolbox) and measuring errors, i.e. a complete classification use case scenario. As different datasets require a different amount of time and memory we decided to normalize the charts. The results on the chart are normalized for each dataset separately. To normalize we calculated the median of every library's time or memory performance and selected the highest of the medians in the dataset as the normalization point, i.e. 100\% is the worst median performance. We present the best, median, and worst performance of each library per dataset, normalized performance with the worst median performance on that dataset. The closer the library is to point 0, the better it performed, thus the smaller area inside the library curve, the better. 

\begin{figure}[!b]
  \includegraphics[width=1\textwidth]{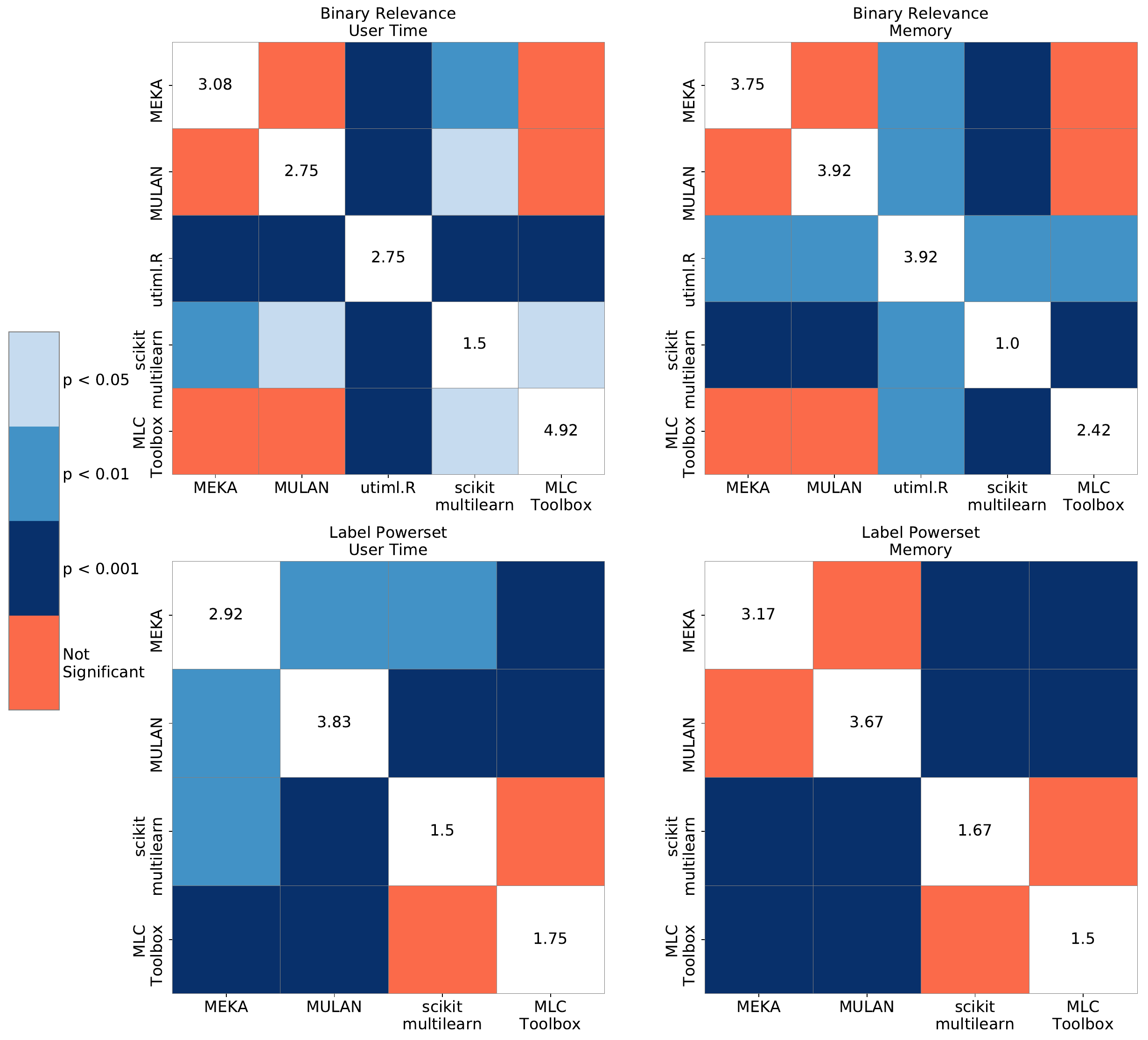}
  \caption{User running time (s) and memory usage ranks of utiml, MLC Toolbox, scikit-multilearn, Meka and Mulan, with RandomForest, with RandomForest, and different multi-label classifiers. There are mean ranks on diagonal.\label{fig:benchmark_statistics}}
\end{figure}

All the libraries were forced to use a single core using the \texttt{taskset} command to minimize parallelization effects on the comparison. Time and memory results were obtained using the \texttt{time -v} command and represent User time, and Maximum resident set size respectively. All results taken into consideration reported that 100\% of their CPU core had been assigned to the process which performed the classification scenario.

We notice that in both classification schemes scikit-multilearn always uses less or, in a few edge cases the same amount of, memory than MEKA or MULAN due to its sparse matrix support. In most cases scikit-multilearn also operates faster than MEKA, MULAN. 

When it comes to label space division approaches modeled in the experiment by Binary Relevance, scikit-multilearn is the most efficient choice on every data set. With Label Powerset multi-class transformation it is in most cases more efficient than MEKA and MULAN and outperforms the MLC Toolbox on datasets with larger numbers of label combinations while on smaller data sets the MLC Toolbox tends to be more efficient. 

We have also compared Binary Relevance from scikit-learn to Binary Relevance from scikit-multilearn. The only difference between the two implementations is the fact that scikit-learn does not convert the output matrix to a Compressed Sparse Column matrix\footnote{\url{https://docs.scipy.org/doc/scipy/reference/generated/scipy.sparse.csc_matrix.html}} while scikit-multilearn does. The observed results were that scikit-learn's implementation was always slightly faster than scikit-multilearn's, while it always used a slightly larger amount of memory, both changes can be attributed to a difference between formats used for the output space. We, therefore, do not report this comparison in extensive detail.

As there is no statistical procedure that allows performing a non-parametric repeated measure tests on multiple measurements per test, we decided to compare the worst case time/memory usage of scikit-multilearn against the best case MEKA/MULAN/utiml/MLC Toolbox performance. We used the Quade test with Bergmann-Hommel post hoc multiple hypothesis testing scenario \citep[as laid out by][]{garcia2010advanced}, the significances of rank difference and mean ranks of each libraries performance are provided in Figure \ref{fig:benchmark_statistics}. 

We see that even in this negative averaging scenario scikit-multilearn is statistically significantly more efficient the other libraries when label space division approach (Binary Relevance) is deployed.

When Label Powerset multi-class transformation is used both scikit-multilearn and MLC Toolbox perform better than MULAN/MEKA with statistical significance. scikit-multilearn ranks slightly higher in speed than the MLC Toolbox but consumes a little more memory. The differences between scikit-multilearn and the MLC Toolbox in this approach are not statistically significant.

\section{Conclusions\label{sec:conc}}
The presented library - scikit-multilearn - is the most extensive scikit-learn compatible multi-label classification library. It provides implementations of both the most popular algorithms and new families of methods such as network-based label space division approaches. It is fast thanks to performing transformations on sparse matrices internally and using well-optimized methods from scikit as base classifiers. scikit-multilearn integrates well with the Python data science stack of scipy. It also provides wrappers to Keras models which allows quick adaptation of deep learning approaches to multi-label classification via problem-transformation approaches, and MEKA/WEKA with parts of MULAN - the standard Java classification stack - and allows easy use of those methods with the rest of the Python stack.

\acks{We would like to thank Wojciech Stachowski for initial versions of BR/ML-kNN \& Classifier Chains implementations, Christian Schulze for asking multiple questions, reporting and fixing bugs, Felipe Almeida for bugfixes and testing, and Fernando Benites for providing the implementation of ML-ARAM \citep{Brucker2011724}. The work was partially supported by The National Science Centre the research projects no. 2016/21/N/ST6/02382 and 2016/21/D/ST6/02948, by the European Union’s Horizon 2020 research and innovation programme under the Marie Skłodowska-Curie grant agreement No 691152 (RENOIR); the Polish Ministry of Science and Higher Education fund for supporting internationally co-financed projects in 2016-2019 (agreement no. 3628/H2020/2016/2) and by the Faculty of Computer Science and Management, Wrocław University of Science and Technology statutory funds.}


\vskip 0.2in
\bibliography{main}

\appendix
\label{app:benchmark}

\begin{table}[p]
\centering
\begin{adjustbox}{max width=.85\textwidth}
\begin{tabular}{llrrrrrrrr}
\toprule
      &         &  count &      mean &     std &       min &       25\% &       50\% &       75\% &       max \\
set & library &        &           &         &           &           &           &           &           \\
\midrule
\multirow{5}{*}{\rot{90}{Corel5k}} & MEKA &     10 &    329.35 &    6.93 &    311.28 &    328.73 &    330.65 &    331.59 &    336.41 \\
      & MLC-Toolbox &     10 &    540.39 &   64.40 &    469.11 &    489.02 &    524.54 &    584.42 &    652.68 \\
      & MULAN &     10 &    323.59 &    5.40 &    316.28 &    321.09 &    322.94 &    326.08 &    332.94 \\
      & scikit-multilearn &     10 &     24.05 &    0.93 &     22.60 &     23.18 &     24.71 &     24.77 &     24.85 \\
      & utiml.R &     10 &   1928.82 &   54.03 &   1825.97 &   1906.33 &   1929.56 &   1961.09 &   1995.39 \\
\cline{1-10}
\multirow{5}{*}{\rot{90}{bibtex}} & MEKA &     10 &    317.91 &    7.41 &    307.67 &    311.13 &    319.67 &    325.21 &    326.59 \\
      & MLC-Toolbox &     10 &    696.19 &   34.06 &    633.80 &    680.56 &    696.30 &    718.40 &    749.24 \\
      & MULAN &     10 &    288.15 &    6.33 &    284.30 &    284.87 &    285.51 &    286.77 &    303.88 \\
      & scikit-multilearn &     10 &     41.05 &    1.74 &     37.19 &     41.17 &     41.81 &     42.00 &     42.38 \\
      & utiml.R &     10 &   1001.93 &   43.17 &    925.99 &   1005.15 &   1007.94 &   1010.66 &   1075.69 \\
\cline{1-10}
\multirow{5}{*}{\rot{90}{birds}} & MEKA &     10 &      3.63 &    0.16 &      3.38 &      3.56 &      3.60 &      3.68 &      3.90 \\
      & MLC-Toolbox &     10 &      1.05 &    0.03 &      1.02 &      1.02 &      1.04 &      1.08 &      1.10 \\
      & MULAN &     10 &      4.30 &    0.10 &      4.09 &      4.27 &      4.34 &      4.38 &      4.40 \\
      & scikit-multilearn &     10 &      1.50 &    0.05 &      1.45 &      1.46 &      1.50 &      1.54 &      1.58 \\
      & utiml.R &     10 &      7.04 &    0.04 &      6.97 &      7.02 &      7.04 &      7.06 &      7.10 \\
\cline{1-10}
\multirow{5}{*}{\rot{90}{delicious}} & MEKA &     10 &   5291.10 &   76.51 &   5193.85 &   5257.89 &   5288.31 &   5298.34 &   5483.73 \\
      & MLC-Toolbox &     10 &  12294.52 &  519.71 &  11082.67 &  12141.24 &  12302.92 &  12510.35 &  13018.59 \\
      & MULAN &     10 &   4902.66 &   37.00 &   4861.75 &   4879.77 &   4896.75 &   4901.54 &   4969.57 \\
      & scikit-multilearn &     10 &    549.79 &   18.06 &    525.48 &    529.89 &    562.26 &    563.51 &    565.50 \\
      & utiml.R &     10 &   6680.74 &  539.44 &   5930.92 &   6453.18 &   6549.90 &   7168.62 &   7381.27 \\
\cline{1-10}
\multirow{5}{*}{\rot{90}{emotions}} & MEKA &     10 &      2.31 &    0.04 &      2.20 &      2.31 &      2.32 &      2.33 &      2.35 \\
      & MLC-Toolbox &     10 &      0.60 &    0.02 &      0.58 &      0.59 &      0.60 &      0.62 &      0.63 \\
      & MULAN &     10 &      2.32 &    0.06 &      2.25 &      2.28 &      2.31 &      2.34 &      2.43 \\
      & scikit-multilearn &     10 &      0.75 &    0.06 &      0.67 &      0.72 &      0.72 &      0.74 &      0.86 \\
      & utiml.R &     10 &      3.89 &    0.03 &      3.84 &      3.89 &      3.89 &      3.91 &      3.93 \\
\cline{1-10}
\multirow{5}{*}{\rot{90}{enron}} & MEKA &     10 &     17.74 &    0.27 &     17.37 &     17.58 &     17.72 &     17.86 &     18.21 \\
      & MLC-Toolbox &     10 &     28.62 &    0.27 &     28.31 &     28.45 &     28.53 &     28.76 &     29.19 \\
      & MULAN &     10 &     15.53 &    0.20 &     15.32 &     15.35 &     15.48 &     15.69 &     15.84 \\
      & scikit-multilearn &     10 &      6.40 &    0.23 &      5.82 &      6.44 &      6.48 &      6.52 &      6.57 \\
      & utiml.R &     10 &     93.75 &    0.55 &     92.72 &     93.34 &     93.99 &     94.13 &     94.27 \\
\cline{1-10}
\multirow{5}{*}{\rot{90}{genbase}} & MEKA &     10 &     11.94 &    0.60 &     11.19 &     11.71 &     11.80 &     12.18 &     13.32 \\
      & MLC-Toolbox &     10 &      2.95 &    0.07 &      2.86 &      2.92 &      2.95 &      3.00 &      3.09 \\
      & MULAN &     10 &      9.69 &    0.21 &      9.39 &      9.56 &      9.68 &      9.74 &     10.15 \\
      & scikit-multilearn &     10 &      5.01 &    0.22 &      4.52 &      4.89 &      5.12 &      5.16 &      5.18 \\
      & utiml.R &     10 &     41.89 &    0.33 &     41.48 &     41.64 &     41.88 &     41.97 &     42.49 \\
\cline{1-10}
\multirow{5}{*}{\rot{90}{mediamill}} & MEKA &     10 &    607.94 &   79.02 &    448.97 &    624.48 &    638.55 &    655.44 &    662.81 \\
      & MLC-Toolbox &     10 &   1446.38 &   73.95 &   1285.85 &   1419.56 &   1463.09 &   1504.47 &   1518.34 \\
      & MULAN &     10 &    424.29 &   23.80 &    415.32 &    416.04 &    416.89 &    417.42 &    491.95 \\
      & scikit-multilearn &     10 &    595.14 &   19.48 &    566.27 &    576.67 &    608.73 &    610.54 &    611.08 \\
      & utiml.R &     10 &   2701.30 &  102.56 &   2536.77 &   2606.49 &   2704.90 &   2800.31 &   2807.30 \\
\cline{1-10}
\multirow{5}{*}{\rot{90}{medical}} & MEKA &     10 &      6.66 &    0.15 &      6.43 &      6.54 &      6.65 &      6.78 &      6.87 \\
      & MLC-Toolbox &     10 &      5.06 &    0.14 &      4.83 &      4.99 &      5.08 &      5.14 &      5.24 \\
      & MULAN &     10 &      7.17 &    0.32 &      6.90 &      7.01 &      7.08 &      7.14 &      8.04 \\
      & scikit-multilearn &     10 &      3.65 &    0.07 &      3.58 &      3.58 &      3.64 &      3.71 &      3.76 \\
      & utiml.R &     10 &     69.57 &    0.19 &     69.35 &     69.44 &     69.59 &     69.63 &     69.99 \\
\cline{1-10}
\multirow{5}{*}{\rot{90}{scene}} & MEKA &     10 &      4.86 &    0.16 &      4.72 &      4.73 &      4.83 &      4.89 &      5.23 \\
      & MLC-Toolbox &     10 &      2.21 &    0.03 &      2.16 &      2.19 &      2.21 &      2.23 &      2.26 \\
      & MULAN &     10 &      4.82 &    0.27 &      4.50 &      4.59 &      4.76 &      5.06 &      5.26 \\
      & scikit-multilearn &     10 &      3.68 &    0.09 &      3.56 &      3.60 &      3.70 &      3.76 &      3.78 \\
      & utiml.R &     10 &     12.47 &    0.10 &     12.27 &     12.43 &     12.48 &     12.55 &     12.59 \\
\cline{1-10}
\multirow{5}{*}{\rot{90}{tmc2007}} & MEKA &     10 &    154.56 &    3.17 &    147.05 &    153.69 &    155.76 &    156.33 &    158.31 \\
      & MLC-Toolbox &     10 &    874.41 &   37.77 &    810.27 &    859.82 &    884.38 &    897.00 &    930.94 \\
      & MULAN &     10 &    146.53 &    5.67 &    143.46 &    143.76 &    143.88 &    144.15 &    157.80 \\
      & scikit-multilearn &     10 &     47.94 &    0.98 &     46.83 &     47.02 &     47.95 &     48.85 &     48.99 \\
      & utiml.R &     10 &   1039.77 &    2.68 &   1036.34 &   1038.47 &   1038.92 &   1041.36 &   1044.98 \\
\cline{1-10}
\multirow{5}{*}{\rot{90}{yeast}} & MEKA &     10 &      6.73 &    0.50 &      6.21 &      6.47 &      6.64 &      6.85 &      8.00 \\
      & MLC-Toolbox &     10 &      5.29 &    0.05 &      5.20 &      5.26 &      5.30 &      5.33 &      5.35 \\
      & MULAN &     10 &      6.10 &    0.14 &      5.91 &      6.05 &      6.09 &      6.14 &      6.43 \\
      & scikit-multilearn &     10 &      4.35 &    0.11 &      4.23 &      4.25 &      4.34 &      4.45 &      4.48 \\
      & utiml.R &     10 &     13.24 &    0.06 &     13.15 &     13.19 &     13.24 &     13.28 &     13.33 \\
\bottomrule
\end{tabular}

\end{adjustbox}
\caption{User time of library running Binary Relevance for each of the datasets.}
\end{table}

\begin{table}[p]
\centering\begin{adjustbox}{max width=.85\textwidth}
\begin{tabular}{llrrrrrrrr}
\toprule
      &         &  count &      mean &     std &       min &       25\% &       50\% &       75\% &       max \\
set & library &        &           &         &           &           &           &           &           \\
\midrule
\multirow{5}{*}{\rot{90}{Corel5k}} & MEKA &     10 &    329.35 &    6.93 &    311.28 &    328.73 &    330.65 &    331.59 &    336.41 \\
      & MLC-Toolbox &     10 &    540.39 &   64.40 &    469.11 &    489.02 &    524.54 &    584.42 &    652.68 \\
      & MULAN &     10 &    323.59 &    5.40 &    316.28 &    321.09 &    322.94 &    326.08 &    332.94 \\
      & scikit-multilearn &     10 &     24.05 &    0.93 &     22.60 &     23.18 &     24.71 &     24.77 &     24.85 \\
      & utiml.R &     10 &   1928.82 &   54.03 &   1825.97 &   1906.33 &   1929.56 &   1961.09 &   1995.39 \\
\cline{1-10}
\multirow{5}{*}{\rot{90}{bibtex}} & MEKA &     10 &    317.91 &    7.41 &    307.67 &    311.13 &    319.67 &    325.21 &    326.59 \\
      & MLC-Toolbox &     10 &    696.19 &   34.06 &    633.80 &    680.56 &    696.30 &    718.40 &    749.24 \\
      & MULAN &     10 &    288.15 &    6.33 &    284.30 &    284.87 &    285.51 &    286.77 &    303.88 \\
      & scikit-multilearn &     10 &     41.05 &    1.74 &     37.19 &     41.17 &     41.81 &     42.00 &     42.38 \\
      & utiml.R &     10 &   1001.93 &   43.17 &    925.99 &   1005.15 &   1007.94 &   1010.66 &   1075.69 \\
\cline{1-10}
\multirow{5}{*}{\rot{90}{birds}} & MEKA &     10 &      3.63 &    0.16 &      3.38 &      3.56 &      3.60 &      3.68 &      3.90 \\
      & MLC-Toolbox &     10 &      1.05 &    0.03 &      1.02 &      1.02 &      1.04 &      1.08 &      1.10 \\
      & MULAN &     10 &      4.30 &    0.10 &      4.09 &      4.27 &      4.34 &      4.38 &      4.40 \\
      & scikit-multilearn &     10 &      1.50 &    0.05 &      1.45 &      1.46 &      1.50 &      1.54 &      1.58 \\
      & utiml.R &     10 &      7.04 &    0.04 &      6.97 &      7.02 &      7.04 &      7.06 &      7.10 \\
\cline{1-10}
\multirow{5}{*}{\rot{90}{delicious}} & MEKA &     10 &   5291.10 &   76.51 &   5193.85 &   5257.89 &   5288.31 &   5298.34 &   5483.73 \\
      & MLC-Toolbox &     10 &  12294.52 &  519.71 &  11082.67 &  12141.24 &  12302.92 &  12510.35 &  13018.59 \\
      & MULAN &     10 &   4902.66 &   37.00 &   4861.75 &   4879.77 &   4896.75 &   4901.54 &   4969.57 \\
      & scikit-multilearn &     10 &    549.79 &   18.06 &    525.48 &    529.89 &    562.26 &    563.51 &    565.50 \\
      & utiml.R &     10 &   6680.74 &  539.44 &   5930.92 &   6453.18 &   6549.90 &   7168.62 &   7381.27 \\
\cline{1-10}
\multirow{5}{*}{\rot{90}{emotions}} & MEKA &     10 &      2.31 &    0.04 &      2.20 &      2.31 &      2.32 &      2.33 &      2.35 \\
      & MLC-Toolbox &     10 &      0.60 &    0.02 &      0.58 &      0.59 &      0.60 &      0.62 &      0.63 \\
      & MULAN &     10 &      2.32 &    0.06 &      2.25 &      2.28 &      2.31 &      2.34 &      2.43 \\
      & scikit-multilearn &     10 &      0.75 &    0.06 &      0.67 &      0.72 &      0.72 &      0.74 &      0.86 \\
      & utiml.R &     10 &      3.89 &    0.03 &      3.84 &      3.89 &      3.89 &      3.91 &      3.93 \\
\cline{1-10}
\multirow{5}{*}{\rot{90}{enron}} & MEKA &     10 &     17.74 &    0.27 &     17.37 &     17.58 &     17.72 &     17.86 &     18.21 \\
      & MLC-Toolbox &     10 &     28.62 &    0.27 &     28.31 &     28.45 &     28.53 &     28.76 &     29.19 \\
      & MULAN &     10 &     15.53 &    0.20 &     15.32 &     15.35 &     15.48 &     15.69 &     15.84 \\
      & scikit-multilearn &     10 &      6.40 &    0.23 &      5.82 &      6.44 &      6.48 &      6.52 &      6.57 \\
      & utiml.R &     10 &     93.75 &    0.55 &     92.72 &     93.34 &     93.99 &     94.13 &     94.27 \\
\cline{1-10}
\multirow{5}{*}{\rot{90}{genbase}} & MEKA &     10 &     11.94 &    0.60 &     11.19 &     11.71 &     11.80 &     12.18 &     13.32 \\
      & MLC-Toolbox &     10 &      2.95 &    0.07 &      2.86 &      2.92 &      2.95 &      3.00 &      3.09 \\
      & MULAN &     10 &      9.69 &    0.21 &      9.39 &      9.56 &      9.68 &      9.74 &     10.15 \\
      & scikit-multilearn &     10 &      5.01 &    0.22 &      4.52 &      4.89 &      5.12 &      5.16 &      5.18 \\
      & utiml.R &     10 &     41.89 &    0.33 &     41.48 &     41.64 &     41.88 &     41.97 &     42.49 \\
\cline{1-10}
\multirow{5}{*}{\rot{90}{mediamill}} & MEKA &     10 &    607.94 &   79.02 &    448.97 &    624.48 &    638.55 &    655.44 &    662.81 \\
      & MLC-Toolbox &     10 &   1446.38 &   73.95 &   1285.85 &   1419.56 &   1463.09 &   1504.47 &   1518.34 \\
      & MULAN &     10 &    424.29 &   23.80 &    415.32 &    416.04 &    416.89 &    417.42 &    491.95 \\
      & scikit-multilearn &     10 &    595.14 &   19.48 &    566.27 &    576.67 &    608.73 &    610.54 &    611.08 \\
      & utiml.R &     10 &   2701.30 &  102.56 &   2536.77 &   2606.49 &   2704.90 &   2800.31 &   2807.30 \\
\cline{1-10}
\multirow{5}{*}{\rot{90}{medical}} & MEKA &     10 &      6.66 &    0.15 &      6.43 &      6.54 &      6.65 &      6.78 &      6.87 \\
      & MLC-Toolbox &     10 &      5.06 &    0.14 &      4.83 &      4.99 &      5.08 &      5.14 &      5.24 \\
      & MULAN &     10 &      7.17 &    0.32 &      6.90 &      7.01 &      7.08 &      7.14 &      8.04 \\
      & scikit-multilearn &     10 &      3.65 &    0.07 &      3.58 &      3.58 &      3.64 &      3.71 &      3.76 \\
      & utiml.R &     10 &     69.57 &    0.19 &     69.35 &     69.44 &     69.59 &     69.63 &     69.99 \\
\cline{1-10}
\multirow{5}{*}{\rot{90}{scene}} & MEKA &     10 &      4.86 &    0.16 &      4.72 &      4.73 &      4.83 &      4.89 &      5.23 \\
      & MLC-Toolbox &     10 &      2.21 &    0.03 &      2.16 &      2.19 &      2.21 &      2.23 &      2.26 \\
      & MULAN &     10 &      4.82 &    0.27 &      4.50 &      4.59 &      4.76 &      5.06 &      5.26 \\
      & scikit-multilearn &     10 &      3.68 &    0.09 &      3.56 &      3.60 &      3.70 &      3.76 &      3.78 \\
      & utiml.R &     10 &     12.47 &    0.10 &     12.27 &     12.43 &     12.48 &     12.55 &     12.59 \\
\cline{1-10}
\multirow{5}{*}{\rot{90}{tmc2007}} & MEKA &     10 &    154.56 &    3.17 &    147.05 &    153.69 &    155.76 &    156.33 &    158.31 \\
      & MLC-Toolbox &     10 &    874.41 &   37.77 &    810.27 &    859.82 &    884.38 &    897.00 &    930.94 \\
      & MULAN &     10 &    146.53 &    5.67 &    143.46 &    143.76 &    143.88 &    144.15 &    157.80 \\
      & scikit-multilearn &     10 &     47.94 &    0.98 &     46.83 &     47.02 &     47.95 &     48.85 &     48.99 \\
      & utiml.R &     10 &   1039.77 &    2.68 &   1036.34 &   1038.47 &   1038.92 &   1041.36 &   1044.98 \\
\cline{1-10}
\multirow{5}{*}{\rot{90}{yeast}} & MEKA &     10 &      6.73 &    0.50 &      6.21 &      6.47 &      6.64 &      6.85 &      8.00 \\
      & MLC-Toolbox &     10 &      5.29 &    0.05 &      5.20 &      5.26 &      5.30 &      5.33 &      5.35 \\
      & MULAN &     10 &      6.10 &    0.14 &      5.91 &      6.05 &      6.09 &      6.14 &      6.43 \\
      & scikit-multilearn &     10 &      4.35 &    0.11 &      4.23 &      4.25 &      4.34 &      4.45 &      4.48 \\
      & utiml.R &     10 &     13.24 &    0.06 &     13.15 &     13.19 &     13.24 &     13.28 &     13.33 \\
\bottomrule
\end{tabular}

\end{adjustbox}
\caption{Maximum memory usage of library running Binary Relevance for each of the datasets.}
\end{table}

\begin{table}[p]
\centering\begin{adjustbox}{max width=.9\textwidth}
\begin{tabular}{llrrrrrrrr}
\toprule
      &                   &  count &     mean &    std &      min &      25\% &      50\% &      75\% &      max \\
set & library &        &          &        &          &          &          &          &          \\
\midrule
\multirow{4}{*}{\rot{90}{Corel5k}} & MEKA &     10 &    47.93 &   1.06 &    46.67 &    47.08 &    47.60 &    48.82 &    49.81 \\
      & MLC-Toolbox &     10 &    17.19 &   0.24 &    16.86 &    16.98 &    17.21 &    17.34 &    17.54 \\
      & MULAN &     10 &    54.92 &   1.88 &    51.15 &    54.24 &    55.19 &    56.07 &    57.34 \\
      & scikit-multilearn &     10 &     8.56 &   0.33 &     8.15 &     8.28 &     8.48 &     8.81 &     9.06 \\
\cline{1-10}
\multirow{4}{*}{\rot{90}{bibtex}} & MEKA &     10 &    46.60 &   1.37 &    43.60 &    45.98 &    46.64 &    47.56 &    48.48 \\
      & MLC-Toolbox &     10 &    28.83 &   0.47 &    28.15 &    28.62 &    28.84 &    29.08 &    29.70 \\
      & MULAN &     10 &    53.81 &   1.66 &    52.24 &    52.62 &    52.81 &    55.21 &    56.44 \\
      & scikit-multilearn &     10 &    26.36 &   0.14 &    26.23 &    26.25 &    26.32 &    26.40 &    26.66 \\
\cline{1-10}
\multirow{4}{*}{\rot{90}{birds}} & MEKA &     10 &     2.03 &   0.17 &     1.82 &     1.88 &     1.98 &     2.21 &     2.25 \\
      & MLC-Toolbox &     10 &     0.40 &   0.01 &     0.38 &     0.40 &     0.40 &     0.40 &     0.41 \\
      & MULAN &     10 &     2.60 &   0.04 &     2.53 &     2.57 &     2.58 &     2.63 &     2.66 \\
      & scikit-multilearn &     10 &     0.85 &   0.05 &     0.77 &     0.84 &     0.85 &     0.86 &     0.97 \\
\cline{1-10}
\multirow{4}{*}{\rot{90}{delicious}} & MEKA &     10 &  1406.20 &  86.09 &  1264.17 &  1357.53 &  1427.58 &  1480.17 &  1486.04 \\
      & MLC-Toolbox &     10 &   131.27 &   4.79 &   123.27 &   128.79 &   130.55 &   135.06 &   137.92 \\
      & MULAN &     10 &  1606.88 &  63.40 &  1478.83 &  1623.74 &  1632.84 &  1642.51 &  1651.81 \\
      & scikit-multilearn &     10 &    59.40 &   1.15 &    57.64 &    58.30 &    59.73 &    60.41 &    60.68 \\
\cline{1-10}
\multirow{4}{*}{\rot{90}{emotions}} & MEKA &     10 &     1.63 &   0.11 &     1.49 &     1.55 &     1.58 &     1.72 &     1.82 \\
      & MLC-Toolbox &     10 &     0.37 &   0.01 &     0.36 &     0.36 &     0.36 &     0.37 &     0.38 \\
      & MULAN &     10 &     2.16 &   0.03 &     2.13 &     2.14 &     2.16 &     2.18 &     2.21 \\
      & scikit-multilearn &     10 &     0.53 &   0.06 &     0.47 &     0.48 &     0.52 &     0.59 &     0.63 \\
\cline{1-10}
\multirow{4}{*}{\rot{90}{enron}} & MEKA &     10 &     7.54 &   0.43 &     7.07 &     7.21 &     7.46 &     7.69 &     8.39 \\
      & MLC-Toolbox &     10 &     1.91 &   0.03 &     1.87 &     1.89 &     1.90 &     1.91 &     1.98 \\
      & MULAN &     10 &     8.39 &   0.26 &     7.94 &     8.22 &     8.36 &     8.64 &     8.71 \\
      & scikit-multilearn &     10 &     3.54 &   0.09 &     3.42 &     3.48 &     3.52 &     3.63 &     3.67 \\
\cline{1-10}
\multirow{4}{*}{\rot{90}{genbase}} & MEKA &     10 &     3.94 &   0.19 &     3.67 &     3.81 &     3.90 &     4.10 &     4.21 \\
      & MLC-Toolbox &     10 &     0.38 &   0.02 &     0.36 &     0.36 &     0.38 &     0.39 &     0.42 \\
      & MULAN &     10 &     4.32 &   0.19 &     3.81 &     4.32 &     4.36 &     4.42 &     4.47 \\
      & scikit-multilearn &     10 &     2.92 &   0.04 &     2.86 &     2.90 &     2.91 &     2.95 &     2.98 \\
\cline{1-10}
\multirow{4}{*}{\rot{90}{mediamill}} & MEKA &     10 &   646.99 &   7.33 &   637.66 &   640.00 &   648.32 &   652.74 &   657.56 \\
      & MLC-Toolbox &     10 &   210.37 &   4.56 &   205.49 &   207.03 &   207.94 &   214.65 &   218.22 \\
      & MULAN &     10 &   647.64 &  26.74 &   573.99 &   650.70 &   652.64 &   658.56 &   670.37 \\
      & scikit-multilearn &     10 &   479.39 &  15.57 &   445.40 &   475.34 &   486.05 &   490.02 &   491.12 \\
\cline{1-10}
\multirow{4}{*}{\rot{90}{medical}} & MEKA &     10 &     2.74 &   0.19 &     2.36 &     2.67 &     2.74 &     2.78 &     3.05 \\
      & MLC-Toolbox &     10 &     0.55 &   0.02 &     0.54 &     0.54 &     0.55 &     0.56 &     0.59 \\
      & MULAN &     10 &     3.68 &   0.18 &     3.21 &     3.70 &     3.72 &     3.76 &     3.84 \\
      & scikit-multilearn &     10 &     2.52 &   0.08 &     2.30 &     2.52 &     2.54 &     2.55 &     2.59 \\
\cline{1-10}
\multirow{4}{*}{\rot{90}{scene}} & MEKA &     10 &     3.36 &   0.14 &     3.11 &     3.31 &     3.37 &     3.44 &     3.58 \\
      & MLC-Toolbox &     10 &     0.74 &   0.02 &     0.72 &     0.73 &     0.74 &     0.75 &     0.78 \\
      & MULAN &     10 &     3.81 &   0.18 &     3.48 &     3.80 &     3.88 &     3.92 &     3.96 \\
      & scikit-multilearn &     10 &     2.96 &   0.05 &     2.86 &     2.92 &     2.98 &     2.99 &     3.01 \\
\cline{1-10}
\multirow{4}{*}{\rot{90}{tmc2007}} & MEKA &     10 &    85.39 &   3.73 &    76.84 &    86.48 &    86.92 &    87.17 &    88.28 \\
      & MLC-Toolbox &     10 &   139.91 &   0.29 &   139.52 &   139.66 &   139.93 &   140.10 &   140.33 \\
      & MULAN &     10 &    89.85 &   3.21 &    80.90 &    90.44 &    90.71 &    90.83 &    92.52 \\
      & scikit-multilearn &     10 &    32.36 &   1.50 &    29.61 &    31.72 &    33.18 &    33.32 &    33.68 \\
\cline{1-10}
\multirow{4}{*}{\rot{90}{yeast}} & MEKA &     10 &     3.76 &   0.14 &     3.47 &     3.73 &     3.78 &     3.87 &     3.90 \\
      & MLC-Toolbox &     10 &     1.18 &   0.02 &     1.14 &     1.17 &     1.18 &     1.19 &     1.21 \\
      & MULAN &     10 &     4.02 &   0.11 &     3.79 &     3.98 &     4.05 &     4.10 &     4.14 \\
      & scikit-multilearn &     10 &     2.11 &   0.10 &     1.93 &     2.06 &     2.08 &     2.16 &     2.27 \\
\bottomrule
\end{tabular}

\end{adjustbox}
\caption{User time of library running Label Powerset for each of the datasets.}
\end{table}

\begin{table}[p]
\centering\begin{adjustbox}{max width=.9\textwidth}
\begin{tabular}{llrrrrrrrr}
\toprule
      &                   &  count &     mean &    std &      min &      25\% &      50\% &      75\% &      max \\
set & library &        &          &        &          &          &          &          &          \\
\midrule
\multirow{4}{*}{\rot{90}{Corel5k}} & MEKA &     10 &    47.93 &   1.06 &    46.67 &    47.08 &    47.60 &    48.82 &    49.81 \\
      & MLC-Toolbox &     10 &    17.19 &   0.24 &    16.86 &    16.98 &    17.21 &    17.34 &    17.54 \\
      & MULAN &     10 &    54.92 &   1.88 &    51.15 &    54.24 &    55.19 &    56.07 &    57.34 \\
      & scikit-multilearn &     10 &     8.56 &   0.33 &     8.15 &     8.28 &     8.48 &     8.81 &     9.06 \\
\cline{1-10}
\multirow{4}{*}{\rot{90}{bibtex}} & MEKA &     10 &    46.60 &   1.37 &    43.60 &    45.98 &    46.64 &    47.56 &    48.48 \\
      & MLC-Toolbox &     10 &    28.83 &   0.47 &    28.15 &    28.62 &    28.84 &    29.08 &    29.70 \\
      & MULAN &     10 &    53.81 &   1.66 &    52.24 &    52.62 &    52.81 &    55.21 &    56.44 \\
      & scikit-multilearn &     10 &    26.36 &   0.14 &    26.23 &    26.25 &    26.32 &    26.40 &    26.66 \\
\cline{1-10}
\multirow{4}{*}{\rot{90}{birds}} & MEKA &     10 &     2.03 &   0.17 &     1.82 &     1.88 &     1.98 &     2.21 &     2.25 \\
      & MLC-Toolbox &     10 &     0.40 &   0.01 &     0.38 &     0.40 &     0.40 &     0.40 &     0.41 \\
      & MULAN &     10 &     2.60 &   0.04 &     2.53 &     2.57 &     2.58 &     2.63 &     2.66 \\
      & scikit-multilearn &     10 &     0.85 &   0.05 &     0.77 &     0.84 &     0.85 &     0.86 &     0.97 \\
\cline{1-10}
\multirow{4}{*}{\rot{90}{delicious}} & MEKA &     10 &  1406.20 &  86.09 &  1264.17 &  1357.53 &  1427.58 &  1480.17 &  1486.04 \\
      & MLC-Toolbox &     10 &   131.27 &   4.79 &   123.27 &   128.79 &   130.55 &   135.06 &   137.92 \\
      & MULAN &     10 &  1606.88 &  63.40 &  1478.83 &  1623.74 &  1632.84 &  1642.51 &  1651.81 \\
      & scikit-multilearn &     10 &    59.40 &   1.15 &    57.64 &    58.30 &    59.73 &    60.41 &    60.68 \\
\cline{1-10}
\multirow{4}{*}{\rot{90}{emotions}} & MEKA &     10 &     1.63 &   0.11 &     1.49 &     1.55 &     1.58 &     1.72 &     1.82 \\
      & MLC-Toolbox &     10 &     0.37 &   0.01 &     0.36 &     0.36 &     0.36 &     0.37 &     0.38 \\
      & MULAN &     10 &     2.16 &   0.03 &     2.13 &     2.14 &     2.16 &     2.18 &     2.21 \\
      & scikit-multilearn &     10 &     0.53 &   0.06 &     0.47 &     0.48 &     0.52 &     0.59 &     0.63 \\
\cline{1-10}
\multirow{4}{*}{\rot{90}{enron}} & MEKA &     10 &     7.54 &   0.43 &     7.07 &     7.21 &     7.46 &     7.69 &     8.39 \\
      & MLC-Toolbox &     10 &     1.91 &   0.03 &     1.87 &     1.89 &     1.90 &     1.91 &     1.98 \\
      & MULAN &     10 &     8.39 &   0.26 &     7.94 &     8.22 &     8.36 &     8.64 &     8.71 \\
      & scikit-multilearn &     10 &     3.54 &   0.09 &     3.42 &     3.48 &     3.52 &     3.63 &     3.67 \\
\cline{1-10}
\multirow{4}{*}{\rot{90}{genbase}} & MEKA &     10 &     3.94 &   0.19 &     3.67 &     3.81 &     3.90 &     4.10 &     4.21 \\
      & MLC-Toolbox &     10 &     0.38 &   0.02 &     0.36 &     0.36 &     0.38 &     0.39 &     0.42 \\
      & MULAN &     10 &     4.32 &   0.19 &     3.81 &     4.32 &     4.36 &     4.42 &     4.47 \\
      & scikit-multilearn &     10 &     2.92 &   0.04 &     2.86 &     2.90 &     2.91 &     2.95 &     2.98 \\
\cline{1-10}
\multirow{4}{*}{\rot{90}{mediamill}} & MEKA &     10 &   646.99 &   7.33 &   637.66 &   640.00 &   648.32 &   652.74 &   657.56 \\
      & MLC-Toolbox &     10 &   210.37 &   4.56 &   205.49 &   207.03 &   207.94 &   214.65 &   218.22 \\
      & MULAN &     10 &   647.64 &  26.74 &   573.99 &   650.70 &   652.64 &   658.56 &   670.37 \\
      & scikit-multilearn &     10 &   479.39 &  15.57 &   445.40 &   475.34 &   486.05 &   490.02 &   491.12 \\
\cline{1-10}
\multirow{4}{*}{\rot{90}{medical}} & MEKA &     10 &     2.74 &   0.19 &     2.36 &     2.67 &     2.74 &     2.78 &     3.05 \\
      & MLC-Toolbox &     10 &     0.55 &   0.02 &     0.54 &     0.54 &     0.55 &     0.56 &     0.59 \\
      & MULAN &     10 &     3.68 &   0.18 &     3.21 &     3.70 &     3.72 &     3.76 &     3.84 \\
      & scikit-multilearn &     10 &     2.52 &   0.08 &     2.30 &     2.52 &     2.54 &     2.55 &     2.59 \\
\cline{1-10}
\multirow{4}{*}{\rot{90}{scene}} & MEKA &     10 &     3.36 &   0.14 &     3.11 &     3.31 &     3.37 &     3.44 &     3.58 \\
      & MLC-Toolbox &     10 &     0.74 &   0.02 &     0.72 &     0.73 &     0.74 &     0.75 &     0.78 \\
      & MULAN &     10 &     3.81 &   0.18 &     3.48 &     3.80 &     3.88 &     3.92 &     3.96 \\
      & scikit-multilearn &     10 &     2.96 &   0.05 &     2.86 &     2.92 &     2.98 &     2.99 &     3.01 \\
\cline{1-10}
\multirow{4}{*}{\rot{90}{tmc2007}} & MEKA &     10 &    85.39 &   3.73 &    76.84 &    86.48 &    86.92 &    87.17 &    88.28 \\
      & MLC-Toolbox &     10 &   139.91 &   0.29 &   139.52 &   139.66 &   139.93 &   140.10 &   140.33 \\
      & MULAN &     10 &    89.85 &   3.21 &    80.90 &    90.44 &    90.71 &    90.83 &    92.52 \\
      & scikit-multilearn &     10 &    32.36 &   1.50 &    29.61 &    31.72 &    33.18 &    33.32 &    33.68 \\
\cline{1-10}
\multirow{4}{*}{\rot{90}{yeast}} & MEKA &     10 &     3.76 &   0.14 &     3.47 &     3.73 &     3.78 &     3.87 &     3.90 \\
      & MLC-Toolbox &     10 &     1.18 &   0.02 &     1.14 &     1.17 &     1.18 &     1.19 &     1.21 \\
      & MULAN &     10 &     4.02 &   0.11 &     3.79 &     3.98 &     4.05 &     4.10 &     4.14 \\
      & scikit-multilearn &     10 &     2.11 &   0.10 &     1.93 &     2.06 &     2.08 &     2.16 &     2.27 \\
\bottomrule
\end{tabular}

\end{adjustbox}
\caption{Maximum memory usage of library running Label Powerset for each of the datasets.}
\end{table}

\end{document}